\documentclass[10pt,twocolumn,letterpaper]{article}

\usepackage{cvpr}
\usepackage{times}
\usepackage{epsfig}
\usepackage{graphicx}
\usepackage{amsmath}
\usepackage{amssymb}
\usepackage{booktabs}
\usepackage{multirow}
\usepackage{amsthm}
\usepackage{xcolor}

\usepackage[breaklinks=true,bookmarks=false]{hyperref} % colorlinks,%pagebackref=true,

\newcommand{\ns}{{\ensuremath\Delta}}
\usepackage{amsmath, amsfonts, amssymb, amsthm}
\usepackage{txfonts} 
\usepackage{verbatim}
\usepackage{url}
\usepackage{color}
\usepackage{algorithm2e}
\usepackage[noend]{algpseudocode}
\usepackage{multirow}

\usepackage{siunitx}
\renewcommand{\tilde}{\widetilde}

\ifx\BlackBox\undefined
\newcommand{\BlackBox}{\rule{1.5ex}{1.5ex}}  % end of proof
\fi
\ifx\proof\undefined
\newenvironment{proof}{\par\noindent{\bf Proof\ }}{\hfill\BlackBox\\[2mm]}
\fi

\newcommand\shortsection[1]{\vspace{6pt}{\noindent\bf #1.}}

\theoremstyle{definition}
\newtheorem{definition}{Definition}[section]

\makeatletter
\def\url@leostyle{%
  \@ifundefined{selectfont}{\def\UrlFont{\sf}}{\def\UrlFont{\small\sffamily}}}
\makeatother
\makeatletter
\def\url@beostyle{%
  \@ifundefined{selectfont}{\def\UrlFont{\sf}}{\def\UrlFont{\scriptsize\sffamily}}}
\makeatother
\cvprfinalcopy % *** Uncomment this line for the final submission

\def\cvprPaperID{40} % *** Enter the CVPR Paper ID here

\begin{document}

\title{One Neuron to Fool Them All}

\author{Anshuman Suri\\
University of Virginia\\
{\tt\small anshuman@virginia.edu}
% For a paper whose authors are all at the same institution,
% omit the following lines up until the closing ``}''.
% Additional authors and addresses can be added with ``\and'',
% just like the second author.
% To save space, use either the email address or home page, not both
\and
David Evans\\
University of Virginia\\
{\tt\small evans@virginia.edu}
}

\maketitle

\begin{abstract}
Despite vast research in adversarial examples, the root causes of model susceptibility are not well understood. Instead of looking at attack-specific robustness, we propose a notion that evaluates the sensitivity of individual neurons in terms of how robust the model's output is to direct perturbations of that neuron's output. Analyzing models from this perspective reveals distinctive characteristics of standard as well as adversarially-trained robust models, and leads to several curious results. In our experiments on CIFAR-10 and ImageNet, we find that attacks using a loss function that targets just a single sensitive neuron find adversarial examples nearly as effectively as ones that target the full model. We analyze the properties of these sensitive neurons to propose a regularization term that can help a model achieve robustness to a variety of different perturbation constraints while maintaining accuracy on natural data distributions. Code for all our experiments is available at \url{https://github.com/iamgroot42/sauron}.
\end{abstract}

\section{Introduction}

Since the discovery of adversarial examples~\cite{carlini2018audio,  gleave2019adversarial,42503}, there have been many attempts at designing defences~\cite{adv_training, zhang2019theoretically}, synthesizing new attacks that break existing defences~\cite{tramer2020adaptive}, and attempts at understanding the underlying phenomena behind this susceptibility~\cite{mahloujifar2019empirically}. Although recent work has helped reduce computational overheads while training for robustness~\cite{moosavi2019robustness,shafahi2019adversarial,wong2019fast}, known methods also incur significant drops in performance on benign data. Moreover, robustness to a specific adversary is not very useful against an adaptive adversary. Simultaneous robustness to different types of $L_p$ bounded perturbations (e.g., $L_\infty$ and $L_1$) is, in fact, impossible to achieve in simple statistical settings~\cite{tramer2019adversarial}. Observations on actual datasets and machine-learning models further reinforce claims of some perturbation types being mutually exclusive.  

Our work is based on the premise that analyzing the inherent robustness properties of a model can help demystify adversarial susceptibility, and perhaps even lead to methods for improving robustness. A few recent works have attempted to identify and remedy `sensitive' neurons that activate more frequently on perturbed inputs~\cite{sehwag2020pruning,zhang2019interpreting, neuron_select}. However, all these works assume a Projected Gradient Descent (PGD)~\cite{adv_training} attack constrained by particular $L_p$ perturbations. Moreover, like most adversarial defences, they suffer drops in overall accuracy. An ideal measure of model robustness should not make strong assumptions about the adversary. Although some recent work research with latent-space perturbations~\cite{creswell2017latentpoison,zhou2019latent} study adversary-agnostic attacks, the perturbations are either unconstrained in the input space, not guaranteed to trigger misclassification, or cannot be directly extended to classification models. 

\shortsection{Contributions}
We propose a new approach to study the inherent robustness of a machine learning model: instead of looking at performance against a given adversarial input (forward direction), we look at the impact of a neuron's output on the model's output to quantify its adversarial sensitivity (reverse direction).
This notion of \textit{neuron sensitivity} allows each neuron's sensitivity to be measured in terms of how much the output value of that neuron can change without changing the model's classification (Section~\ref{sec:sense_defn}). Since this definition relates to a neuron's output, it is independent of particular attacks, and thus, has advantages over previous definitions based on limiting input perturbations. Our experimental results with single-neuron attacks (Section~\ref{sec:fooling}) show that the proposed sensitivity measure is related to the adversarial susceptibility of a model.

Further, we propose a regularization term to minimize neuron sensitivity~(Section~\ref{sec:robust_train}) and show that we can train a robust model using this additional loss term, without compromising on accuracy on clean data. Models trained with the proposed loss term are not only simultaneously robust to multiple perturbation families, but incur minimal computational overhead: training with this term is nearly as fast as regular training. This does not result in state-of-the-art robustness but suggests that our notion of sensitivity correlates with adversarial susceptibility, and approaches that analyze and train for robustness at the level of individual neurons are a promising direction for both deepening our understanding and generating more robust models.

\section{Neuron Sensitivity} \label{sec:sense_defn}

Consider a deep neural network, $D$, designed for a classification problem. For a neuron identified by $i$, we define the \emph{sensitivity} for that neuron for an input $x$ as:

\theoremstyle{definition}
\begin{definition}[Neuron Sensitivity]
We define the \emph{sensitivity}, $\ns$, of neuron $i$ for an input $x$ as the minimum perturbation to the neuron's output that results in a misclassification:
\begin{align}
\label{eq:delta_defn}
\begin{split}
    \ns(i, x) = \operatorname*{argmin}_{\delta}
         \{\; |\delta| \mid D(x) \neq \tilde{D}^{i}_{\delta}(x)\}
\end{split}
\end{align}
\noindent
where $\tilde{D}^{i}_\delta$ corresponds to a modified network that adds perturbation $\delta$ to the output of neuron $i$.
\end{definition}

We posit that this notion of sensitivity can help capture the inherent robustness of a model and not just robustness against specific adversaries.  Throughout the paper, we refer to these sensitivities as $\ns$-values: a higher $\ns$-value implies lower sensitivity, as a larger perturbation is required on that neuron's output for misclassification. Since an adversarial input is supposed to cause considerable changes in the model's output for a slightly perturbed input, these changes must stem from internal outputs of the model itself, which we aim to capture via our definition of sensitivity. We use a summary statistic for sensitivities to define an ordering over neurons. Additionally, we can use these sensitivities to craft adversarial inputs (Section~\ref{sec:fooling}).

\subsection{Identifying Sensitive Neurons}

Analyzing the sensitivity of neurons corresponding to the logits layer is a good starting point since it has an easy closed-form solution. The inputs to the logits layer can be considered as feature representations, $z(x)$, learned by the model~\cite{ilyas2019adversarial}. Let the class of a given input $x$ be $y$, $l(x)$ be the corresponding logits, and $W$ be the logits-layer weight matrix. Then, the sensitivity of neuron $i$ for input $x$ is: 
\begin{align}
    \ns(i,x) =  \dfrac{l(x)[y] - l(x)[\hat{y}]}{W[i, \hat{y}] - W[i, y]} \label{eq:delta_solution}
\end{align}

\begin{figure}
\begin{center}
    \includegraphics[width=1.0\linewidth]{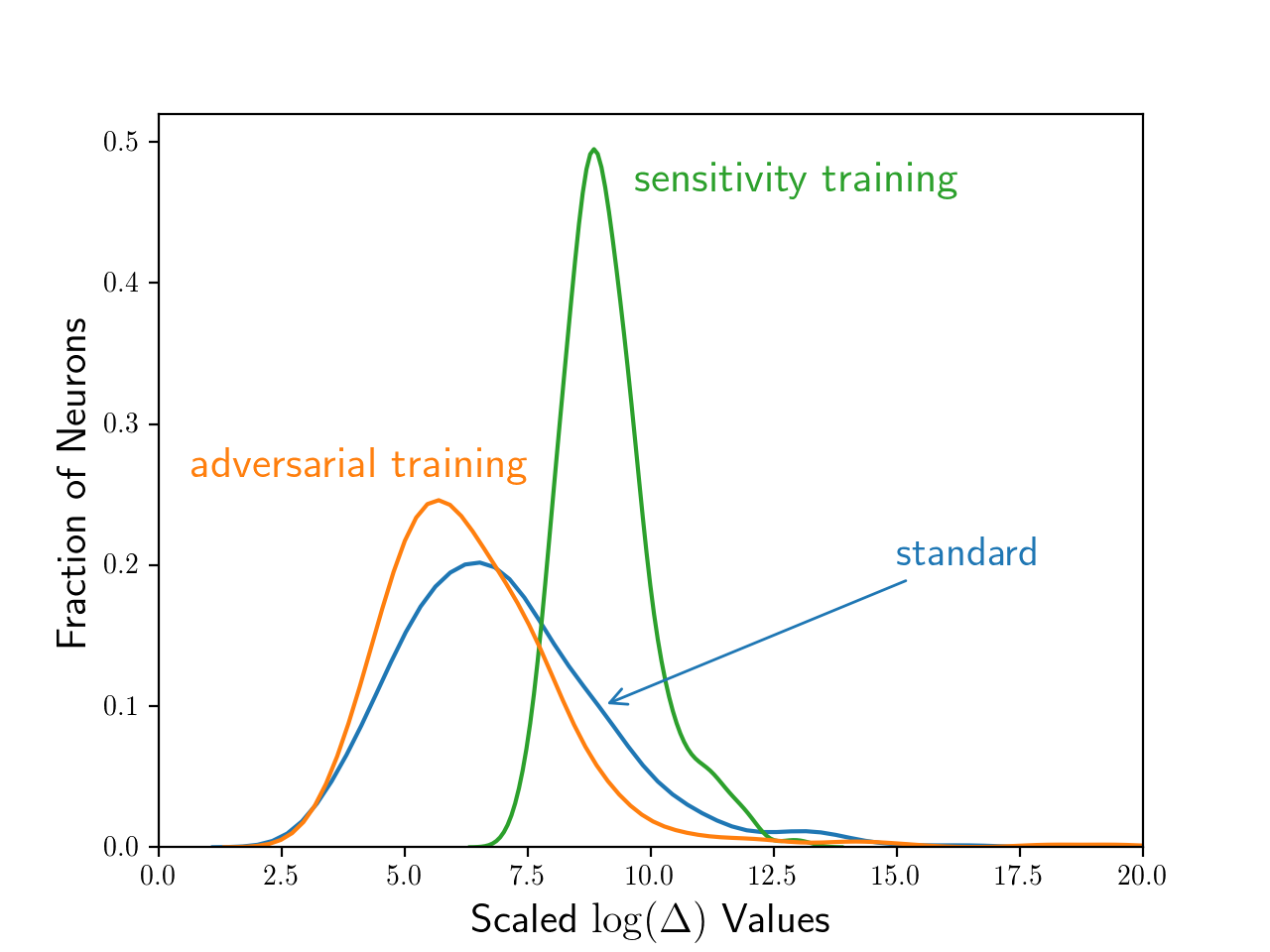}
\end{center}
\caption{Distribution of scaled neuron sensitivities for VGG-19 models trained normally (blue), with adversarial training using an $L_\infty$ PGD adversary (orange), and with the proposed regularization term (green) on CIFAR-10. Histograms are smoothed with a Gaussian Kernel, with Scott Kernel-size estimation.}
\label{fig:delta_values}
\end{figure}

\noindent Interpreting inputs to the logits layer as features learned by the model, these sensitivity values have an interesting interpretation as a measure inversely related to how robust the model's output is to slight changes in specific features. In an ideal scenario, no single feature should have too much influence over the model's classification output. The ideal classifier should learn redundant features that are independent yet highly correlated, given the input's actual class. Such a solution is intuitive as well: tires in an airplane can easily fool a classifier for CIFAR-10~\cite{krizhevsky2009learning} that learns to give very high importance to the presence of tires for vehicles, or eyes that resemble tires. Focusing on the sensitivity of neurons is a step towards achieving this goal of truly-robust learning.

As an illustration of the potential value of our sensitivity definition, Figure~\ref{fig:delta_values} shows the distribution of  $\ns$-values over the neurons for three different VGG-19~\cite{simonyan2014very} models. Both the standard and adversarially-trained models follow similar distributions: the majority of neurons have moderate $\ns$-values, and most of them are quite low, which can be exploited by an adversary since it would target the most sensitive neurons. In Section~\ref{sec:robust_train}, we introduce a training regularization term motivated by our sensitivity definition that produces the ``sensitivity-trained'' model, with a distribution where nearly all neurons have high $\ns$-values.

\subsection{Neuron Sensitivity Attacks} \label{sec:fooling}

Since it is possible to compute $\ns$-values for neurons in the last layer, we consider methods for finding adversarial examples considering only a single neuron. We consider an adversary that wishes to maximize the attack success rate (adversary successfully changes model's prediction) for the inputs that they craft. Moreover, empirically measuring attack success rates for neurons with different $\ns$-values can help establish a correlation between neuron sensitivity and adversarial susceptibility. 
Using the sensitivity of a neuron $i$ in the logits layer for input $x$, we can define a loss function to find adversarial inputs $x^{'}$ that minimizes:
\begin{align} \label{eq:find_adv_seed}
    \left\lVert D(x^{'}) - \tilde{D}^{i}_{\ns(i,x)}(x)\right\rVert_2
\end{align}
\noindent
We then solve for $x^{'}$ with constrained optimization: using a Lagrangian multiplier formulation with gradient descent. The adversary may even use PGD to incorporate perturbation budgets in the input space. Since this attack is easily parallelizable across target neurons, it is feasible for the adversary to target multiple features in the logits layer. 

\begin{figure}
\begin{center}
    \includegraphics[width=1.0\linewidth]{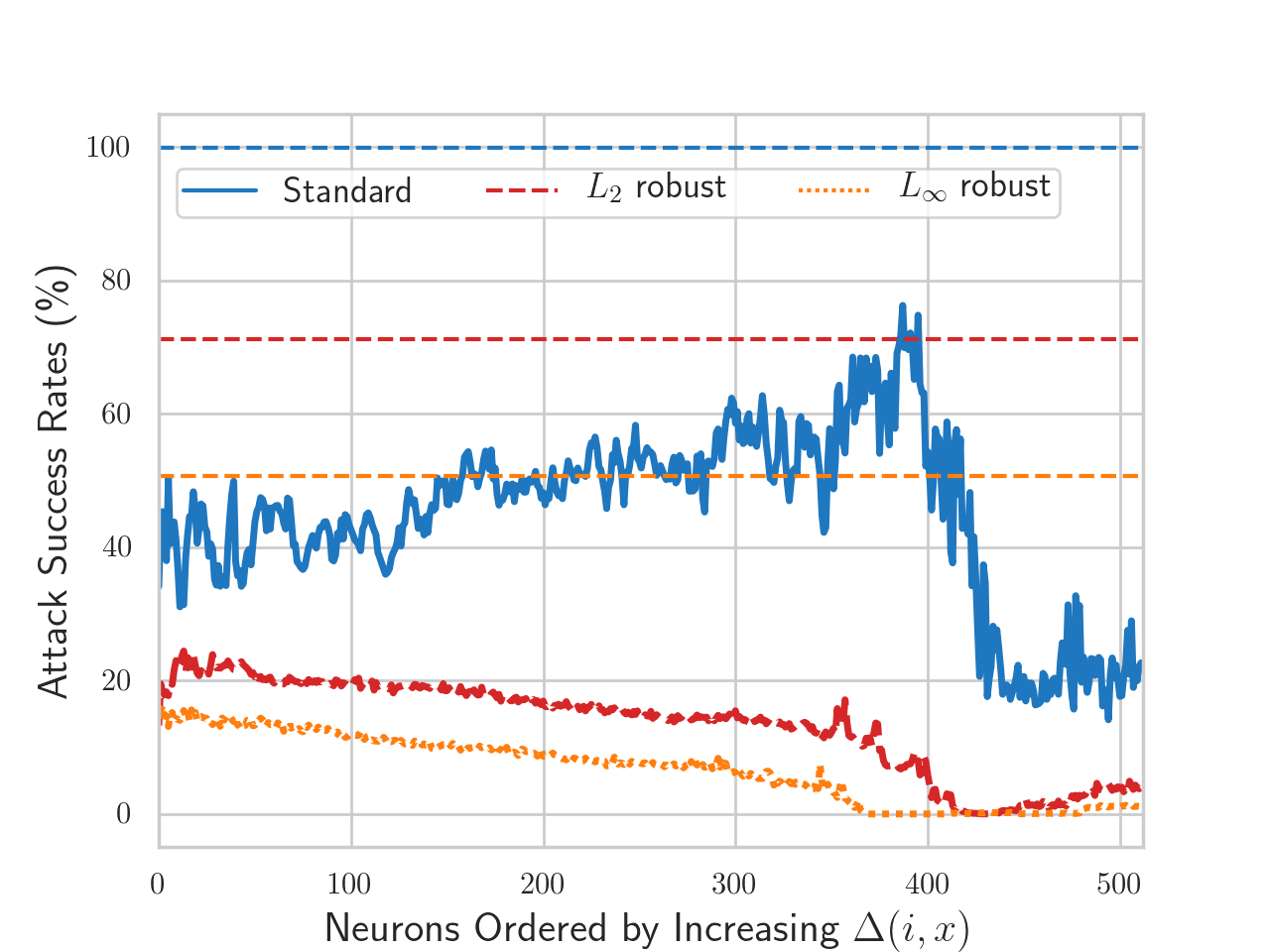}
\end{center}
  \caption{Attack success rates for single-neuron attacks targeting different neurons, ordered by increasing $\ns(i,x)$ for each input seed $x$, for an $L_\infty (\epsilon=\frac{8}{255})$ constrained adversary.  Dashed lines represent attack success rates using standard PGD attack.}
\label{fig:attack_rate_trend}
\end{figure}

Through our experiments, we observe interesting (and sometimes surprising) correlations between the sensitivity-based ordering of neurons and success rates of the proposed attack, as shown in  Figure~\ref{fig:attack_rate_trend}. Each index on the x-axis represents picking the $i^{th}$ neuron for each input $x$, when ordered by $\ns(i,x)$. 
For both the adversarially-trained robust models, there is a range where we see a strong correlation of the relationship we expected: attack success rates (percentage of seeds for which adversary finds a successful adversarial example) drop with increasing $\ns$-values. Beyond a certain point ($\sim$400 out of 512 neurons), the attack success rates stop following any comprehensible patterns. This is because those neurons correspond to features with close to zero variance throughout the dataset. Thus, by definition, such neurons have $\ns$-values that are unlikely to be reached by any input. Nonetheless, using those $\ns$-values as an objective helps the adversary successfully find examples that trigger misclassifications.

To further improve attacks, an adversary can use multiple ($n$, hyper-parameter) target neurons and run the single-neuron attack on each one of them independently. If any one of these succeeds, we count this as a successful $n$-neuron attack. As evident from Table~\ref{tab:attack_numbers}, even a single neuron attack achieves non-trivial attack success rates against all models and trying the top 16 neurons approaches the success rate of full PGD. It is not surprising that this attack is not as successful as PGD: it targets only one neuron at a time and is thus not expected to be not as powerful. Nonetheless, the fact that targeting just a handful of neurons works is useful for finding adversarial examples seems surprising. 

\section{Training for Minimal Sensitivity} \label{sec:robust_train}

\begin{table*}
\begin{center}
\begin{tabular}{cc|c|ccccc|ccccc}
\toprule
% & & PGD $L_2$ & \multicolumn{4}{|c|}{Ours $L_2$} & PGD $L_\infty$ & \multicolumn{4}{c|}{Ours $L_\infty$} \\ \midrule
 & \multicolumn{1}{c}{} & \multicolumn{1}{c}{Normal} & \multicolumn{5}{c}{$L_2$ Attack success rates} & \multicolumn{5}{c}{$L_\infty$ Attack success rates} \\
 & \multicolumn{1}{c}{Model} & \multicolumn{1}{c}{Accuracy} & \multicolumn{1}{c}{PGD} & 1-NS & 4-NS & 8-NS & 16-NS & \multicolumn{1}{c}{PGD} & 1-NS & 4-NS & 8-NS & 16-NS\\ \midrule
\multirow{4}{*}{
\rotatebox[origin=c]{90}{CIFAR-10}} & Standard & 93.0 & 97.8 & 33.0 & 71.7 & 93.1 & 97.7 & 99.9 & 34.1 & 74.9 & 96.6 & 99.9 \\ % \midrule
& $L_2$-Robust & 86.7 & 34.8 & 9.4 & 23.0 & 28.4 & 31.8 & 71.2 & 13.3 & 38.4 & 50.8 & 66.7 \\ 
& $L_\infty$-Robust & 78.3 & 34.4   & 12.5 & 24.4 & 29.8 & 33.1 & 50.7 & 15.7 & 33.2 & 41.2 & 47.9 \\ % \midrule
& Sensitivity & 92.9 & 40.8 & 15.1 & 36.0 & 42.7 & 45.5 & 49.5 & 27.4 & 56.9 & 64.6 & 66.9\\ 
\midrule 
\multirow{3}{*}{
\rotatebox[origin=c]{90}{ImageNet}} & Standard & 76.1 & 99.9 & 96.1 & 99.9 & 99.9 & 99.9 & 99.9 & 98.1 & 100 & 100 & 100\\
& $L_2$ Robust & 57.9 & 58.3 & 30.4 & 48.1 & 46.0 & 47.3 & 66.5 & 34.9 & 48.1 & 50.8 & 52.4\\ 
& $L_\infty$ Robust & 62.4 & 85.8 & 53.3 & 68.3 & 71.9 & 74.3 & 62.3 & 31.9 & 44.4 & 47.4 &  48.6\\ [3pt] \bottomrule
\end{tabular}
\end{center}
\caption{Accuracies and attack success rates using PGD (100 iterations, 20 restarts) and neuron sensitivity attacks for $L_2$ ($\epsilon=0.5$ for CIFAR10, $\epsilon=3.0$ for ImageNet) and $L_\infty$ ($\epsilon=\frac{8}{255}$ for CIFAR10, $\frac{4}{255}$ for ImageNet) perturbation constraints, for VGG-19 trained on CIFAR10 and Resnet-50~\cite{he2016deep} trained on Imagenet~\cite{deng2009imagenet}. The $k$-NS attack is the attack independently targeting the top-$k$ neurons, ranked by $\ns$-sensitivity.}
\label{tab:attack_numbers}
\end{table*}

Our initial motivation for this work was the belief that it might be possible to prune sensitive neurons to help with robustness. However, we found that there is no amount of pruning that improves robustness while maintaining any classification accuracy. Pruning just 20\% of the most sensitive (lowest average $\ns(i, x)$ over training-set inputs) features leads to 9\% and 3\% absolute drops in accuracy and robustness, respectively.  We observed this trend for standard models as well as models trained with an adversarial loss. However, pruning neurons with high $\ns$-values does not impact robustness or accuracy significantly: dropping out nearly 50\% of the model's features makes nearly no difference (which could be explained by redundancy~\cite{ayinde2019correlation}).
These observations are in contradiction to the view of `robust' and `useful' features proposed by Madry~\etal ~\cite{ilyas2019adversarial}. Since an ad-hoc ``patch" to help with robustness does not seem to help, we consider the alternative of constructing a loss function around the sensitivity of neurons.

As single-neuron attack success rates seem correlated with $\ns$-values, one possible way to help with robustness against such attacks is to maximize the minimum $\ns$-value across neurons. Since a simple solution for $\ns$-values exists only for the last layer, a straightforward form of such a regularization term is:

\begin{align} 
    \dfrac{1}{n}\sum^{n}_{i=1}\dfrac{1}{|Y|-1}\sum_{\hat{y} \neq y_{i}} \max_{j \in F}\biggl\lvert\frac{\ns(j, x_i) - \mu_j}{\sigma_j}\biggr\rvert \label{eq:vanilla_reg}
\end{align}
\noindent where $\mu_j$ and $\sigma_j$ are the mean and standard deviation of $z_j$, $|Y|$ is the number of classes, and $F$ is the set of features. However, directly using this loss can lead to degenerate solutions since the model tends to just blow up the logits or make weight values extremely small in order to minimize this objective. Hence, we use a reduced version of the sensitivity regularization term:

\begin{align}
\label{eq:proposed_reg}
\begin{split}
    \dfrac{\lambda_1}{k(|Y| - 1)}\sum_{j}^{\max_{k}}\sum_{l\neq i} |w[j, y_i] - w[j, y_l]| \\
    + \dfrac{\lambda_2}{k}\sum_j^{\max_k}\Bigl\lvert\dfrac{z(x_i)[j](w[j, y_i] - w[j, \hat{y}])}{\sum z(x_i)}\Bigr\rvert
\end{split}
\end{align}
\noindent
The first term encourages the model not to give very high importance to any specific feature for a particular class. The second term encourages the model to ensure that no single feature has a high relative contribution to an input's corresponding logits. 
Summation over the top $k$ neurons helps gradients flow through more neurons, compared to focusing on just the maximum.  Although using either of these terms individually gives non-trivial robustness ($\sim$60\%  $L_2$ PGD attack success rate), combining them has a substantial positive effect, lowering attack success rates significantly. We wish to understand further the complementary nature of these two terms in future work.

We incorporate (\ref{eq:proposed_reg}) in the training optimization procedure. Since this is just a regularization term, it is nearly as fast as regular training. Our experiments show that training with this additional term results in a model that is simultaneously robust to multiple perturbations. Moreover, it does not compromise accuracy on normal inputs (Table~\ref{tab:attack_numbers}). Interestingly, our model has higher robustness against an $L_{\infty}$ adversary than a model trained with an adversarial loss exclusively against the $L_{\infty}$ adversary. The proposed loss term is supposed to lower neuron sensitivity which, as we empirically show in our experiments, is correlated with robustness. Training with the proposed additional loss does not produce state-of-the-art robustness. Nonetheless, it is encouraging as it (a) does not lead to a drop in accuracy, and (b) provides multi-adversary robustness.

\begin{figure}
\begin{center}
    \includegraphics[width=1.0\linewidth]{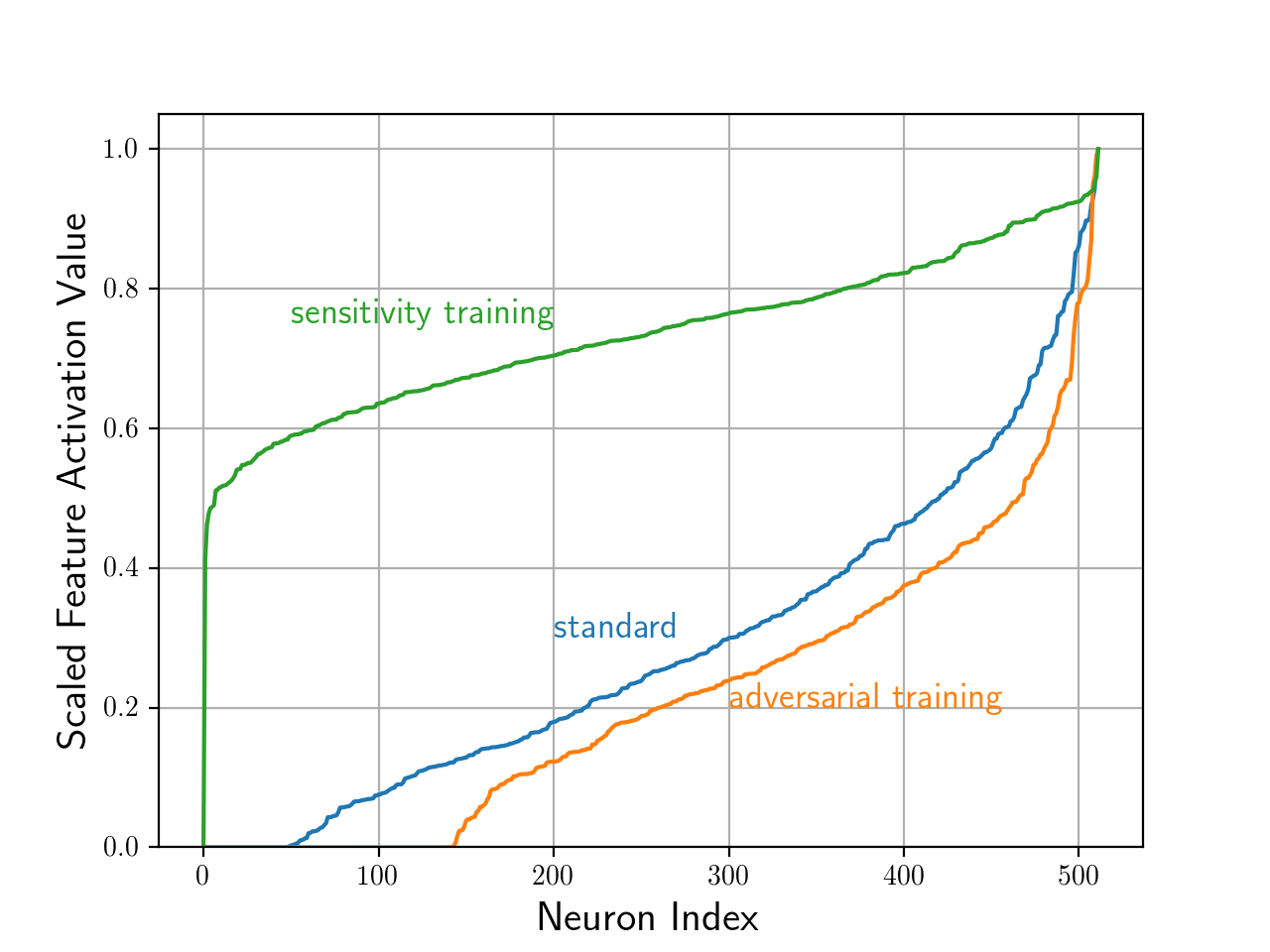}
\end{center}
\caption{Scaled feature activation (min-max scaling) values for VGG-19 models trained normally (blue), with adversarial training (orange), and with the proposed regularization term (green) on CIFAR-10.}
\label{fig:act_values}
\end{figure}

In addition to our robustness results, we observe encouraging trends in the weight and feature distributions of the model. As visible in Figure~\ref{fig:act_values}, adversarial training zeroes out more feature activations than regular training. On the other hand, training for sensitivity leads to nearly all neurons having non-zero activations. Multiple possible explanations could potentially explain this phenomenon, and we are optimistic that further work in this direction will yield improvements in the general understanding of adversarial robustness and insights for better training methods.

\section{Conclusions}

We propose a new way to analyze the inherent robustness of a DNN model. Our experiments show that studying the sensitivity of neurons in a way that is agnostic to input perturbations can be useful for understanding adversarial robustness. Our notion of neuron sensitivity has interesting correlations with adversarial robustness. Additionally, it can be used to find adversarial examples with attack success rates close to PGD attacks. Additionally, a model trained with a regularization to control neuron sensitivity helps with robustness to multiple types of perturbations, without compromising on performance on benign inputs.

We observe a very peculiar phenomenon in our model: using a standard model as a source model to perform black-box attacks yields high success rates, whereas starting with adversarially trained models (or reversing the source-target direction) does not. We suspect this may be because of some kind of decision-boundary similarities between our model and a standard model. We plan on further investigating these observations, along with extending our sensitivity analyses to other layers of the model, as part of future and ongoing work. 

\subsection*{Availability}
\noindent
Open source code for our implementation and for reproducing our experiments is available at: \url{https://github.com/iamgroot42/sauron}.

\subsection*{Acknowledgements}
\noindent
We thank Sicheng Zhu and Xiao Zhang, along with the rest of the Security Research Group and members of the Center for Trustworthy Machine Learning, for insightful discussions and helpful advice on implementing our experiments. This research was sponsored in part by the National Science Foundation \#1804603 (SaTC Frontier: End-to-End Trustworthiness of Machine-Learning Systems), and additional support from Amazon, Baidu, and Intel.

%{\small
\bibliographystyle{ieee_fullname}
\bibliography{references}
%}

\end{document}